%% file: main.tex
\documentclass[10pt,twocolumn,letterpaper]{article}
\usepackage[pagenumbers]{cvpr}

\input{en_preamble}

\definecolor{cvprblue}{rgb}{0.21,0.49,0.74}
\usepackage[unicode,breaklinks,colorlinks,allcolors=cvprblue]{hyperref}
\hypersetup{pdftitle={\modelname{}: Building, Steering, and Memorizing Worlds with Code}}
\hypersetup{pdfauthor={Xuanyi Liu, Ruiqi Li, Siwei Ma}}

\def\paperID{}
\def\confName{CVPR}
\def\confYear{2026}
\title{\modelname{}: Building, Steering, and Memorizing Worlds with Coding Agent}
\author{Xuanyi Liu\thanks{Equal contribution.} \qquad Ruiqi Li\footnotemark[1] \qquad Siwei Ma\\
Peking University}
\date{}

\begin{document}
\maketitle
\input{secs/0_abstract}
\input{secs/1_introduction}
\input{figures/mallea-method-figure-en}
\input{secs/2_related_work}
\input{secs/3_method}
\input{secs/4_experiments}
\input{secs/5_conclusion}

\clearpage
{\small
\bibliographystyle{ieeenat_fullname}
\bibliography{cn_refs}
}
\end{document}

%% file: en_preamble.tex
\usepackage[T1]{fontenc}
\newcommand{\modelname}{Mallea}

%% file: secs/0_abstract.tex
\begin{abstract}
Code world models represent world dynamics through explicit state and executable rules. We investigate how to coordinate object representations, behavioral revisions, and visual memory within this formulation. We present \modelname{}, a system for building, steering, and memorizing worlds from text. A coding agent infers object interaction requirements from the prompt and jointly plans object structure, properties, and rules, favoring compact representations that support those requirements. The world evolves under the planned rules; optional user actions steer it, while text requests can edit properties or extend capabilities with unaffected state preserved. Object texture memory retains reusable appearance only for whitebox-defined objects and adapts it as those objects change. The program advances world state, while a generative renderer combines current structure with remembered appearance. By organizing construction, intervention, and memory around the same objects, \modelname{} conditions subsequent observations on the latest state and compatible appearance.
Project page: \url{https://lxycopper.github.io/Mallea}.
\end{abstract}

%% file: secs/1_introduction.tex
\section{Introduction}
\label{sec:introduction}

Video world models learn visual dynamics from video and simulate scene evolution through generated observations~\cite{hu2023gaia1,yang2023unisim,bruce2024genie,alonso2024diamond,valevski2025gamengen,xiao2025worldmem,zhang2025matrixgame,he2025matrixgame2,sun2025worldplay,robbyant2026lingbot,gao2026lingbot2}. Long-horizon studies report improvements from visual memory, alongside remaining failures of object persistence and physical consistency~\cite{xiao2025worldmem,sun2025worldplay,gao2026lingbot2}. Code world models expose simulated state and executable transition rules, making modeled outcomes accessible to programmatic inspection and revision~\cite{tang2024worldcoder,dainese2024codeworldmodels,chen2026codeworldmodel}. Recent systems connect executable worlds to generative renderers through geometric proxies or registered world surfaces~\cite{chen2026codeworldmodel,huang2026programmable,fang2026coder}. This separates state persistence from rendering fidelity; benchmarks show that visual quality metrics alone do not reliably indicate agreement with prescribed states and events~\cite{huang2026prowbench}.

\input{figures/mallea-positioning-figure-en}

Building, steering, and memorizing a world from text raises three linked questions: how to represent objects, define and change their behavior, and remember their appearance over time. \modelname{} addresses these questions through a coding agent's ability to construct and revise an executable whitebox (Fig.~\ref{fig:positioning}). From the objects and interaction requirements inferred from text, the agent plans representations, manages attributes and rules, and specifies how appearance memory follows object changes. Shared object definitions connect these three designs, allowing the agent's knowledge of the world to guide both simulation and visual memory. Runtime code advances the world, with optional user actions and text requests for steering or revision (Fig.~\ref{fig:mallea-method}).

Our adaptive construction uses the agent's ability to generate object-specific code to seek the simplest representation that preserves required interactions. Following the principle of retaining task-relevant properties in an abstraction~\cite{li2006abstraction}, the agent plans geometry, attributes, and rules together, refining the representation only when an interaction requires additional detail (Fig.~\ref{fig:construction-detail}). Touch and collision may need only coarse contact geometry, whereas strikes and parries can require articulated motion and precise timing. Visual details that do not affect these interactions remain for the generative renderer~\cite{chen2026codeworldmodel,huang2026programmable}. The optics and tactical examples (Fig.~\ref{fig:qualitative-results}) illustrate this separation: mechanical geometry and coarse spatial proxies leave the renderer room to supply scene-specific appearance.

The agent gives these representations behavior through programmable attributes: fields record object properties, and executable rules determine how objects respond. Building on executable rule construction and revision~\cite{tang2024worldcoder,chen2026codeworldmodel,huang2026programmable}, this mechanism offers two connected benefits. \emph{i) Defining behavior per object.} The agent selects properties and rules for each object's interactions, making its responses explicit and individually adjustable, as illustrated by the hammer example (Fig.~\ref{fig:attributes-detail}a). \emph{ii) Extending behavior at runtime.} These explicit definitions let the agent interpret new requests against the current model, then plan, generate, and check missing fields and rules. For example, adding magnetic-response fields and attraction rules lets a world that already supports transporting iron and wood model their different responses to a magnet (Fig.~\ref{fig:attributes-detail}b). Validated revisions preserve existing objects and unrelated state. The agent can therefore revise both current property values and the behavioral capabilities of an existing world. A rendered magnetic-interaction example is included in Fig.~\ref{fig:qualitative-results}.

The agent's control over object structure and behavior also gives it a basis for managing appearance over time. As interactions and edits accumulate over long rollouts, visual consistency involves retaining unchanged details while reflecting changes in the world. Prior work preserves initial-frame anchors, including through attention sinks, or compresses history into bounded context and recurrent states~\cite{yang2025longlive,henschel2025streamingt2v,zhang2025framepack,po2025ssmworld}. Retrieval methods recover distant frames, latents, or KV entries beyond the active window~\cite{xiao2025worldmem,yu2025contextmemory,sun2025worldplay,hu2026longliverag,zhao2026omnimem,zhang2026mosaichunk}. Other methods use 3D geometry to project past appearance into the current view~\cite{wu2025spatialmemory,huang2026programmable,fang2026coder}, or select historical frames that match the objects' current states~\cite{yang2026oneira}. Our object texture memory lets the coding agent manage appearance through the objects it already constructs and edits (Fig.~\ref{fig:memory-mechanism}). It stores textures only for these whitebox objects. The agent uses their attributes and rules to decide what to retain or update, and revises these decisions when new properties or behaviors are introduced. Revisit benchmark results will be released soon.

Our contributions are as follows:
\begin{itemize}[leftmargin=*,nosep]
    \item \textbf{Adaptive object representations.} The coding agent selects compact representations for individual objects to support the interactions inferred from text, leaving other visual details to the renderer.
    \item \textbf{Programmable object behavior.} The agent defines object-specific attributes and rules and extends them during execution in response to new requests.
    \item \textbf{Object texture memory.} The agent manages appearance reuse by object identity and current state, revising reuse conditions as object definitions change.
\end{itemize}
Quantitative benchmark results will be released soon.

%% file: figures/mallea-positioning-figure-en.tex
\begin{figure}[t]
    \centering
    \includegraphics[width=\linewidth]{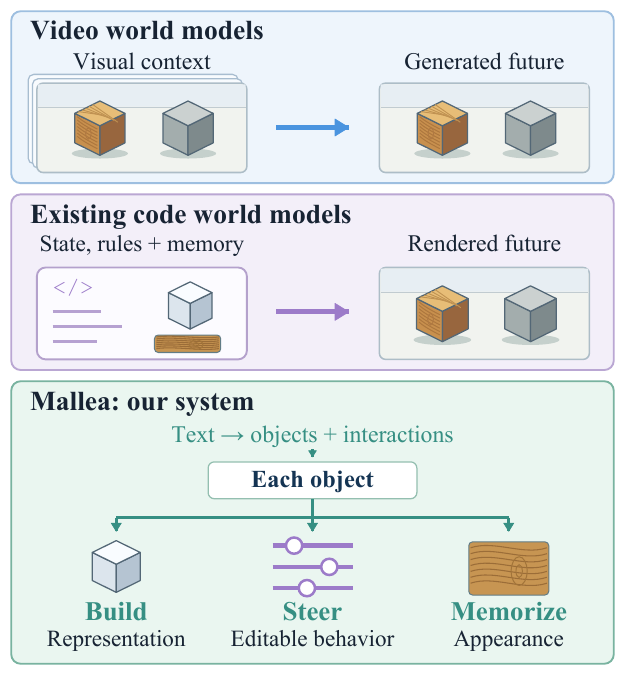}
    \caption{\textbf{Three world-model designs.} \emph{Top:} video models generate future frames from visual context. \emph{Middle:} recent code models use explicit state, rules, and visual memory to guide rendering. \emph{Bottom:} a coding agent in \modelname{} chooses each object's representation (Build), edits its properties and rules (Steer), and manages its stored appearance (Memorize).}
    \label{fig:positioning}
\end{figure}

%% file: figures/mallea-method-figure-en.tex
\begin{figure*}[t]
    \centering
    \includegraphics[width=\textwidth]{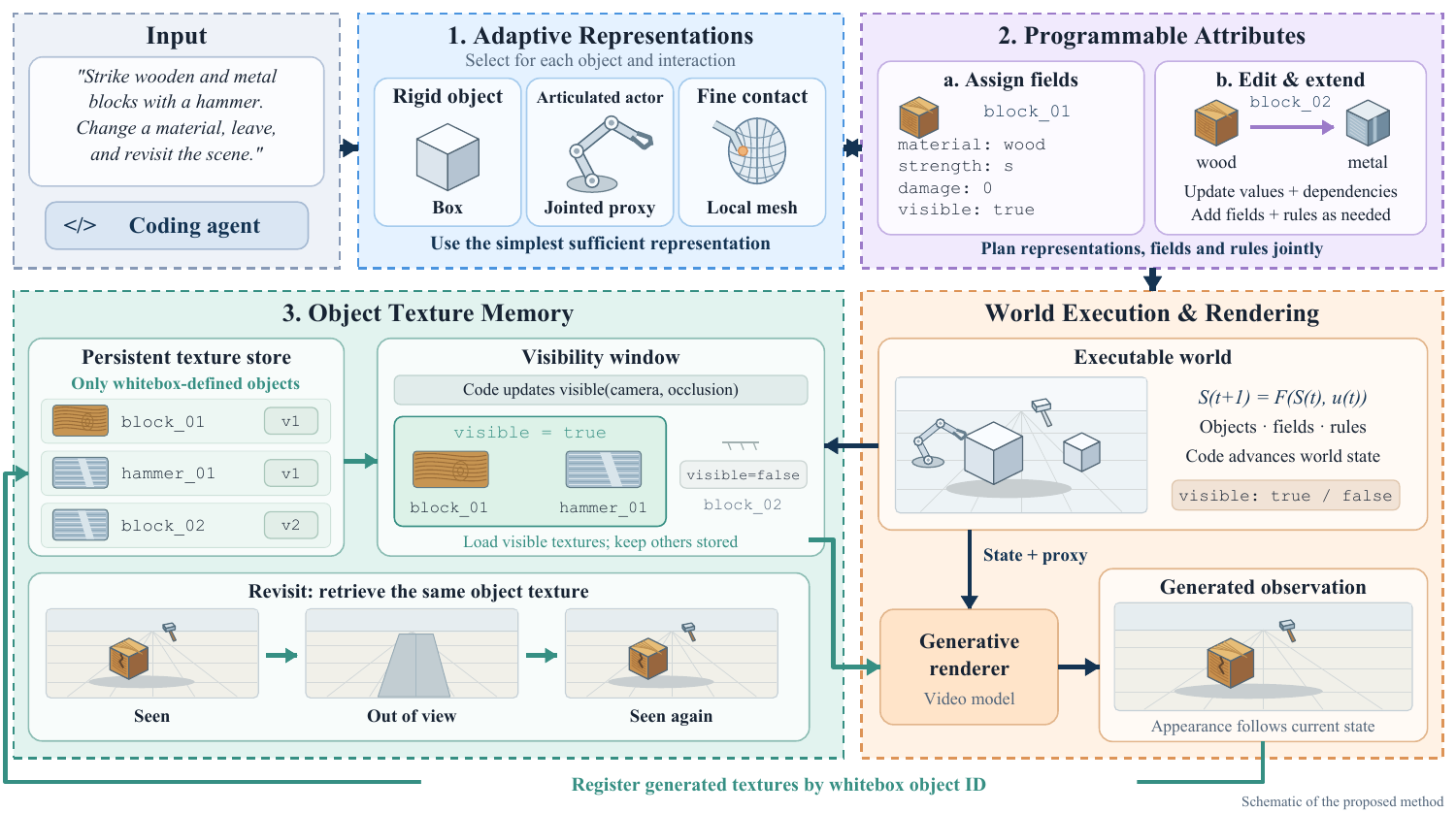}
    \caption{\textbf{Overview of the \modelname{} system.} A coding agent jointly plans representations, fields, and rules. Users may supply action controls or interleave prompts to edit or extend the world. Runtime code advances state and retrieves object textures through a visibility window; generated observations update the store for subsequent revisits. The scenes are schematic illustrations.}
    \label{fig:mallea-method}
\end{figure*}

%% file: secs/2_related_work.tex
\section{Related Work}
\label{sec:related-work}

\textbf{World models through video generation.} Video world models generate visual trajectories, with action-conditioned variants extending generation to user-directed exploration~\cite{bruce2024genie,parkerholder2025genie3,he2025matrixgame2,robbyant2026lingbot,gao2026lingbot2}. Even with visual memory, visually stable rollouts can still fail to preserve object identity or physical consistency~\cite{gao2026lingbot2}. Remembering an observation also differs from exposing the properties and rules that determine an object's next response. \modelname{} maintains interaction-relevant state and rules in executable programs, making state transitions inspectable and enabling targeted interventions.

\textbf{Programmatic modeling and generative rendering.} Language-model agents generate transition programs from action-observation histories~\cite{tang2024worldcoder,dainese2024codeworldmodels}, infer executable physical models from visual evidence~\cite{liang2026visphyworld,wang2026codeasworlds}, and organize scene structure and object interaction logic~\cite{li2026recursive,wu2026code2games}. A complementary line combines explicit state or simulator outputs with generative rendering, allowing structured dynamics to guide richer visual observations~\cite{lin2026stateplay,cai2026mass,meng2026marionette,lv2023gpt4motion,zhan2026magpie}. The division raises a representation question: which information must the structured model retain for the renderer to depict a valid interaction? CWM studies minimally sufficient proxies, while PWM uses state-augmented boxes to balance control and visual detail~\cite{chen2026codeworldmodel,huang2026programmable}. Building on these ideas, \modelname{} jointly selects representations, properties, and rules for individual objects according to the interactions inferred from text. This directs modeling complexity toward behaviorally relevant structure while leaving other visual details to the renderer.

\textbf{Long-horizon consistency and memory.} Memory mechanisms differ in what they retain and how they access it. Attention sinks preserve fixed initial context in streaming models~\cite{xiao2024streamingllm,yang2025longlive}, while compressed history and recurrent states maintain bounded summaries~\cite{zhang2025framepack,po2025ssmworld}. Retrieval methods access frames, latents, or KV entries beyond the local window~\cite{xiao2025worldmem,yu2025contextmemory,hu2026longliverag,zhao2026omnimem}, including fine-grained sections of historical chunks~\cite{zhang2026mosaichunk}; spatial memory additionally reprojects past observations~\cite{wu2025spatialmemory}. Recent CWMs combine visual reuse with executable state: PWM uses temporal and geometry-aligned memories, CoDeR registers appearance on world surfaces, and Oneira filters historical frames by current object state~\cite{huang2026programmable,fang2026coder,yang2026oneira}. These are precedents for persistent appearance and state-aware reuse. \modelname{} organizes texture memory around its adaptively represented, editable whitebox objects, so behavioral revisions determine which appearance remains usable. This connects continued generation to evolving world state and limits the scope of memory updates, with revisits serving as one evaluation of long-horizon consistency.

%% file: secs/3_method.tex
\section{Method}
\label{sec:method}

Mallea connects world construction, object interaction, and appearance memory through shared object identities (Fig.~\ref{fig:mallea-method}). The coding agent builds and revises world code and memory policies; runtime code executes them; a generative renderer produces observations.

\input{figures/mallea-construction-figure-en}
\input{figures/mallea-attributes-figure-en}
\input{figures/mallea-memory-figure-en}
\input{figures/mallea-qualitative-figure-en}

\subsection{Inputs, Outputs, and World State}
\label{sec:world-setup}

The input is an initial text prompt, with optional user actions $u_t$ (e.g., WASD movement) and text edits $q_t$ during execution. The output is a sequence of visual observations $I_t$. Interactions are effects between whitebox objects; actions are user controls. We distinguish the executable model $\mathcal M_t$, which defines objects and rules, from its current state $S_t$, which records their values and relations. Each object has a stable identity, a geometric representation, and typed attributes. Between edits, runtime code advances the world at a fixed simulation step:
\begin{equation}
\label{eq:world}
S_{t+1}=F_{\mathcal M_t}(S_t,u_t).
\end{equation}
Here $u_t=\varnothing$ denotes the absence of a user action; runtime code still advances the world and resolves object interactions from its current geometry, attributes, and rules. An object texture store $\mathcal T_t$ supplies appearance evidence for rendering. It is separate from physical state: leaving the camera view changes retrieval, while the object continues to exist and evolve.

\subsection{Adaptive World Construction}
\label{sec:adaptive}

The agent first identifies the objects and their interactions implied by the prompt, then selects the simplest representations that support them~\cite{chen2026codeworldmodel,huang2026programmable}. As Fig.~\ref{fig:construction-detail} illustrates, coarse collision bounds may suffice for impacts, articulated proxies expose joint motion, and local meshes provide finer contact geometry. These are complementary choices, rather than compulsory stages of refinement.

For each object, the agent plans its representation, attributes, and rules together. A local interaction check asks whether the geometry can supply the contacts and motions required by those rules. Failed checks trigger targeted refinement; successful checks retain the simpler representation. Appearance details that do not affect interaction remain for the renderer.

\subsection{Programmable Attributes}
\label{sec:attributes}

\subsubsection{Defining Behavior per Object}

The agent defines each object's behavioral schema by tracing its planned interactions to the information that response rules read and change (Fig.~\ref{fig:attributes-detail}a). It selects the corresponding fields jointly with the object's geometry and rules. A field specifies its type, unit where applicable, initial value, and permitted writers. Initial values come from the prompt, available evidence, or an explicit modeling assumption. Typed fields also appear in MASS~\cite{cai2026mass}; here field selection is coupled to interaction planning and representation choice.

Rules combine an event, a condition, and an update. For a hammer strike, geometry supplies contact, while material, stiffness, and damage determine the programmed response. Changing a field therefore changes execution. The illustrative responses in the figure assume specified impact and material parameters; a material name alone does not determine a physical outcome. Field checks and contact checks jointly ensure that a rule has both the information and the geometry it requires.

\subsubsection{Extending Behavior at Runtime}

A text request can change an existing value or require new fields and rules. We express an accepted edit as
\begin{equation}
\label{eq:intervention}
(\mathcal M_t^+,S_t^+)=\operatorname{Edit}(\mathcal M_t,S_t,q_t),
\end{equation}
where $+$ denotes the committed revision. A value edit preserves the model definition; an extension revises the definition and initializes its new fields. Both update necessary dependencies and preserve unrelated state.

Figure~\ref{fig:attributes-detail}b shows a world that initially supports transporting iron sheets and wooden blocks. A later request adds magnetic-response fields and attraction rules. Iron responds to the magnet while wood does not respond in the same way. Existing identities, positions, and damage are carried into the extended world.

Before commitment, the agent checks the candidate on a state copy: field types and units, rule dependencies, preservation of unaffected facts, and relevant local behavior. A successful revision is applied at a simulation-step boundary after rechecking its version; failure retains the original world. Unsupported physics requires a validated solver extension. This separates adding a capability from merely invoking an existing one.

\subsection{Object Texture Memory}
\label{sec:memory}

The same object identities index $\mathcal T_t$, connecting current world state to appearance accumulated across long rollouts (Fig.~\ref{fig:memory-mechanism}). Each entry holds reusable appearance, valid surface coverage, and the representation and appearance-related state under which it was registered. Only whitebox-defined objects receive entries; incidental generated background is not automatically stored. Texture coordinates follow box faces, articulated parts, or mesh surfaces, so memory adapts to the chosen representation.

\paragraph{Retrieve for the current view and state.}
The coding agent supplies visibility rules, and runtime code evaluates projection and occlusion to update each object's \texttt{visible} field. The active window contains visible objects; off-screen entries remain stored. For a video segment, retrieval covers objects visible at any point along its committed camera trajectory.

The agent defines reuse conditions from each object's representation and appearance-related attributes, revising them with model extensions. Runtime checks invalidate regions affected by material or damage edits; pose changes can retain object-local textures. Representation changes require a verified surface mapping or invalidate affected texture. View-dependent patches require compatible viewing directions. Compatible regions are projected at current poses to form a masked color condition; unseen, incompatible, or unreliably aligned regions remain unconstrained.

\paragraph{Generate and register appearance.}
The renderer combines the current rendering prompt $p_t$, whitebox-derived geometric conditions $B_t$, and projected appearance $A_t$:
\begin{equation}
\label{eq:memory-rendering}
I_t=G(p_t,B_t,A_t).
\end{equation}
Here, $B_t$ includes the whitebox view and its selected depth or edge controls; $p_t$ reflects the latest world state. The renderer must accept appearance conditioning as well as geometry: depth and edges alone cannot carry stored color.

Following observation-to-surface registration~\cite{fang2026coder}, object IDs, depth, and camera geometry associate the generated observation with whitebox surfaces. Only reliable correspondences update texture coverage; ambiguous matches are rejected. New coverage supplements existing entries, while valid observations refresh regions invalidated by edits. Stored texture is visual evidence, not a guaranteed estimate of intrinsic reflectance.

Reusing texture does not restore a previous world state: the vase retains its pattern while the edited block gains a new appearance (Fig.~\ref{fig:memory-mechanism}). Per-object storage budgets and retrieval over active objects limit texture payload; actual efficiency and consistency require measurement.

Edits affecting a pending rendering segment trigger trajectory and condition recomputation. Evaluation checks physical outcomes against program records~\cite{huang2026prowbench} and tests long-horizon appearance consistency under state changes and delayed revisits.

%% file: figures/mallea-construction-figure-en.tex
\begin{figure}[t]
    \centering
    \includegraphics[width=\linewidth]{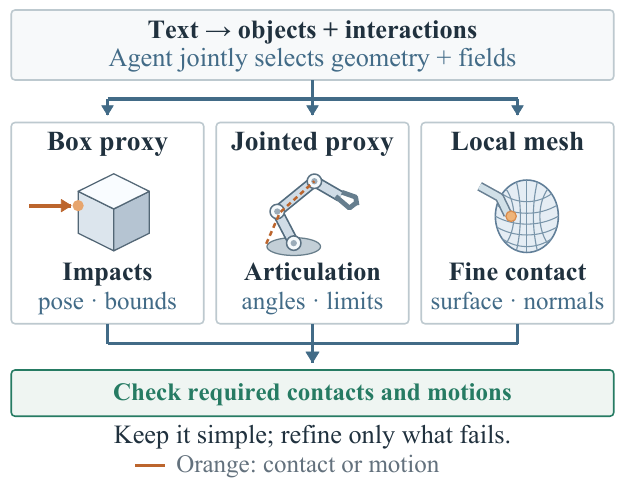}
    \caption{\textbf{Interaction-driven construction.} The agent selects geometry and fields together, checks required contacts and motions, and refines only missing structure. The illustrated representations can be combined.}
    \label{fig:construction-detail}
\end{figure}

%% file: figures/mallea-attributes-figure-en.tex
\begin{figure}[t]
    \centering
    \includegraphics[width=\linewidth]{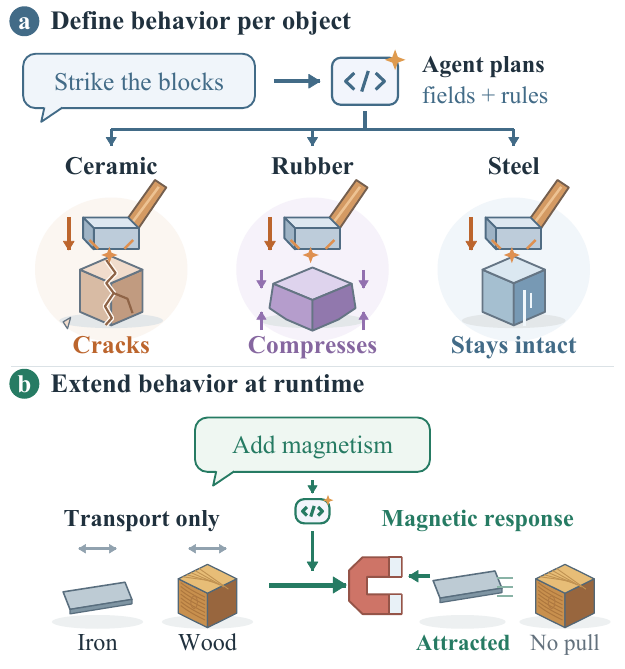}
    \caption{\textbf{Defining and extending behavior.} Code icons denote the agent. (a) Planned fields and rules produce different responses to the same strike under the illustrated material settings. Orange bursts mark contact. (b) A request adds magnetic response to existing iron and wood, preserving unrelated state. Green arrows mark extension and attraction. Mechanism illustrations.}
    \label{fig:attributes-detail}
\end{figure}

%% file: figures/mallea-memory-figure-en.tex
\begin{figure*}[t]
    \centering
    \includegraphics[width=\textwidth]{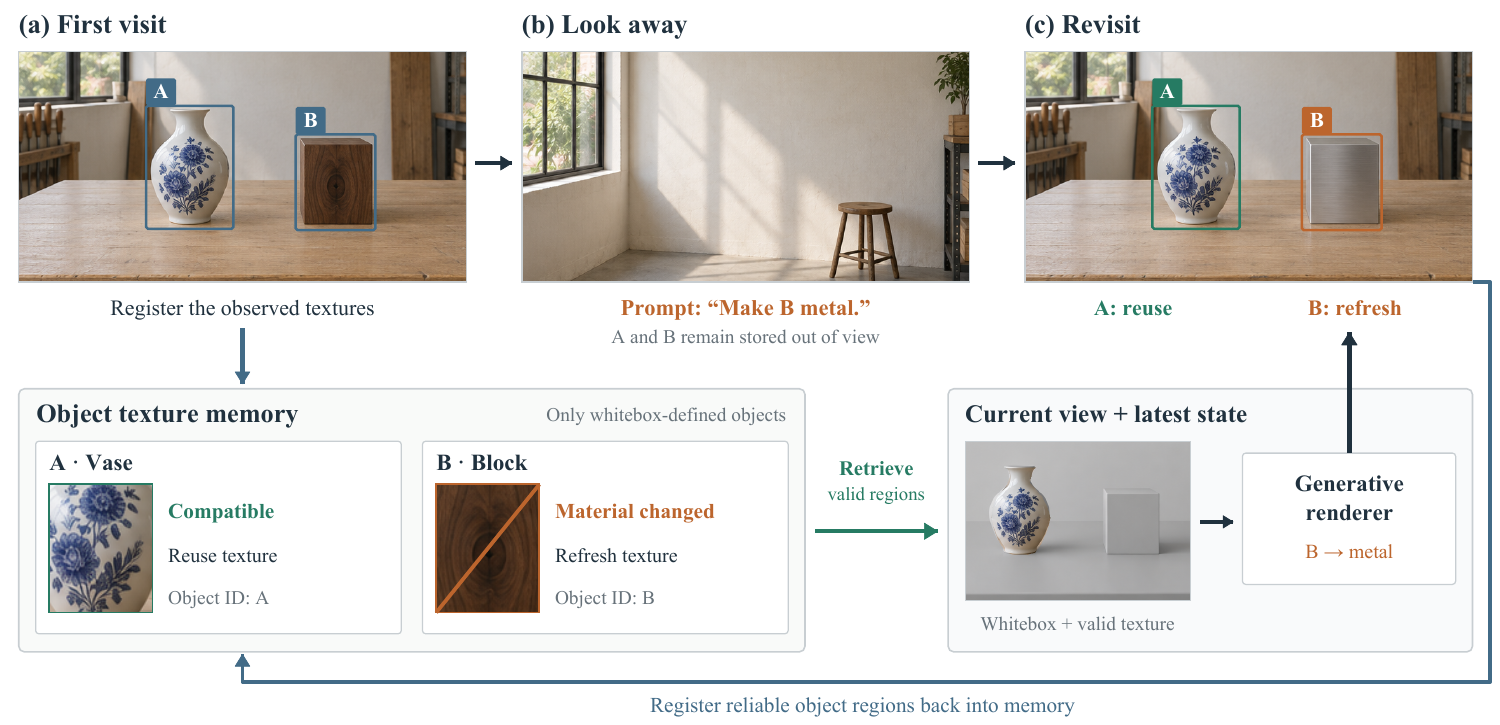}
    \caption{\textbf{Object texture memory across edits and revisits.} The first visit registers textures for whitebox-defined objects A and B. While both are out of view, a text edit changes B's material. On return, the current view and state select compatible appearance: A's pattern is reused (green), while B's invalid texture is refreshed (orange). Generation combines current geometry with valid appearance, then registers reliable object regions back into memory. Off-screen objects retain their entries and continue evolving. Scenes are AI-generated mechanism illustrations, not experimental results.}
    \label{fig:memory-mechanism}
\end{figure*}

%% file: figures/mallea-qualitative-figure-en.tex
\begin{figure*}[t]
    \centering
    \includegraphics[width=\textwidth]{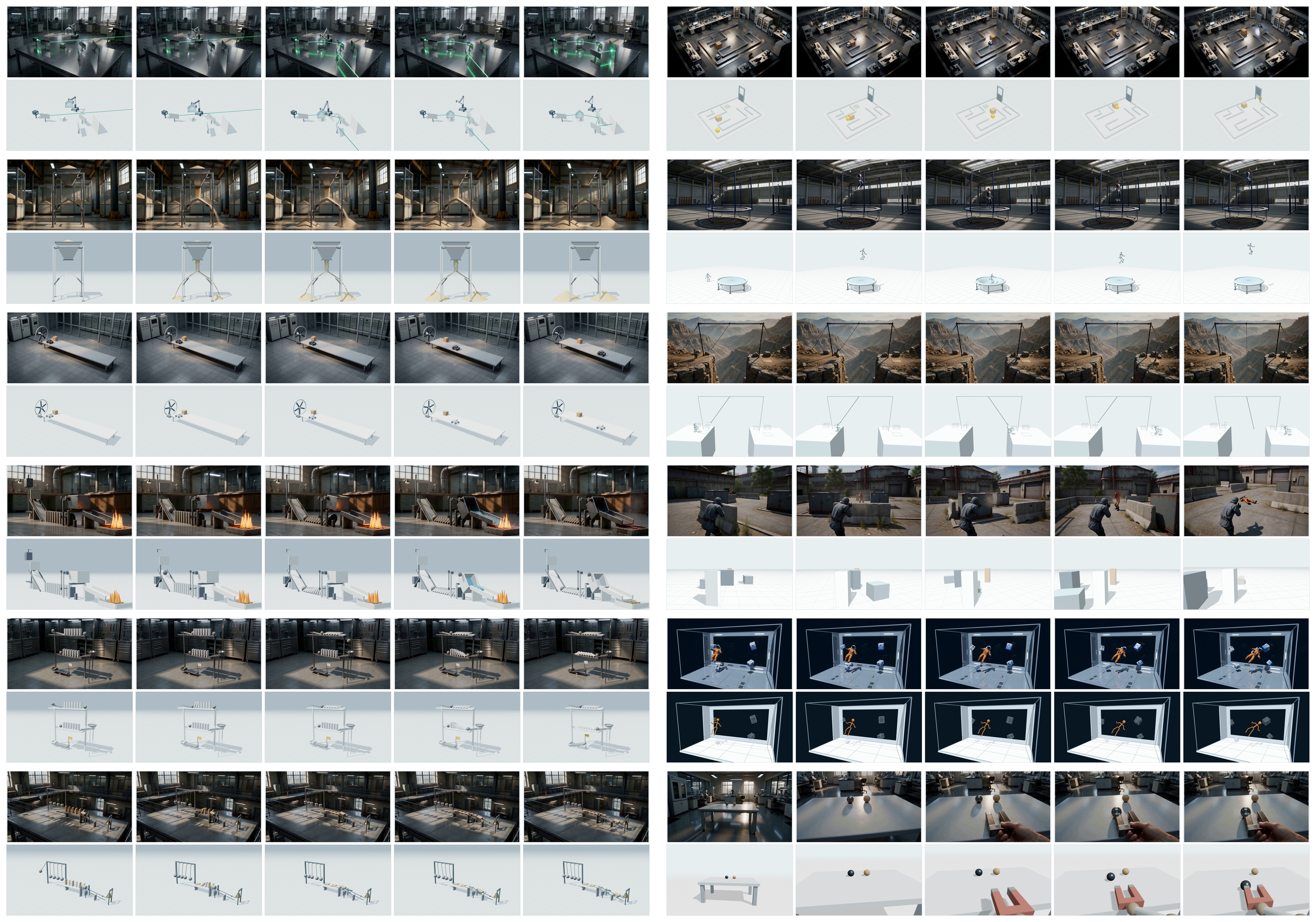}
    \caption{\textbf{Qualitative results.} Generated video frames (top) and corresponding display whitebox frames (bottom) for 12 scenes. Each sequence is sampled at 0\%, 25\%, 50\%, 75\%, and 100\% of its frame span. Sampling is normalized independently, without event alignment. Whiteboxes in left-column rows 2--4 were adjusted after video generation.}
    \label{fig:qualitative-results}
\end{figure*}

%% file: secs/4_experiments.tex
\section{Experiments}
\label{sec:experiments}

\subsection{Qualitative Results}
\label{sec:qualitative-results}

Figure~\ref{fig:qualitative-results} presents 12 scenes spanning physical processes and character motion.

\paragraph{Adaptive representations.}
Mechanical geometry in the optics setup (left, row 1) contrasts with coarse proxies in the tactical scene (right, row 4), while both renderings add scene-specific appearance.

\paragraph{Programmable attributes.}
The fire-control chain (left, row 4) couples object responses; the magnetic scene (right, row 6) illustrates material-dependent behavior. Explicit properties and rules allow these responses to be specified and extended.

\paragraph{Object texture memory.}
The zero-gravity scene (right, row 5) motivates retaining object appearance under pose changes. Our design addresses this through state-aware texture reuse; these short clips do not isolate memory's contribution. Quantitative results, including memory ablations and long-horizon revisit benchmarks, will be released soon.

%% file: secs/5_conclusion.tex
\section{Conclusion}
\label{sec:conclusion}

We presented \modelname{}, a coding-agent system for adaptive object representations, programmable attributes, and object texture memory. Shared executable objects connect interaction geometry, extensible behavior, and state-compatible appearance reuse. Revisions guide execution and rendering while preserving unaffected state. Qualitative examples illustrate varied representations and interactions; quantitative evaluation will assess interaction fidelity, runtime extension, long-horizon consistency, and computational cost.